# Flexible and Approximate Computation through State-Space Reduction

**Weixiong Zhang**
Information Sciences Institute and
Department of Computer Science
University of Southern California
4676 Admiralty Way
Marina del Rey, CA 90292
E-mail: zhang@isi.edu

## Abstract

In the real world, insufficient information, limited computation resources, and complex problem structures often force an autonomous agent to make a decision in time less than that required to solve the problem at hand completely. Flexible and approximate computation are two approaches to decision making under limited computation resources. Flexible computation helps an agent to flexibly allocate limited computation resources so that the overall system utility is maximized. Approximate computation enables an agent to find the best satisfactory solution within a deadline. In this paper, we present two state-space reduction methods for flexible and approximate computation: quantitative reduction to deal with inaccurate heuristic information, and structural reduction to handle complex problem structures. These two methods can be applied successively to continuously improve solution quality if more computation is available. Our results show that these reduction methods are effective and efficient, finding better solutions with less computation than some existing well-known methods.

## 1 Introduction

An autonomous agent must overcome two major difficulties in the real world. First, the available information may not be sufficient or precise to allow the agent to make an optimal decision. Therefore, the agent has to depend upon the best approximate solution to the problem at hand. Second, the available computation resources, especially the computation time, are limited, which may force the agent to make a decision within a deadline.

In situation where computation resources are limited and a deadline must be met, flexible computation can assist an agent to allocate its computation resources in such a way that its overall performance or utility is maximized [5, 9, 14, 32]. The performance of the agent depends not only on the quality of its decision, but also on the amount of computation resources that it uses and the penalty introduced by response delay. This area of research is becoming more and more important and has drawn much attention recently [1, 2, 3]. An important issue of flexible computation is to find the relationship between deliberation, which is the process of searching for a high-quality decision, and the payoff of such a decision [5, 9, 14, 32]. This relationship is usually represented by a performance profile [32]. If the agent has a performance profile of a problem to be solved, it can estimate the amount of computation that it needs in order to find a solution with a satisfactory quality, or vice versa. However, the amount of time allocated for reasoning is generally not known a priori for most applications. Thus, the key to flexible computation is to construct anytime algorithms, which will be described in the next paragraph.

Flexible computation is also closely related to and sometimes relies on approximation methods. Limited computation resources generally prohibit finding optimal solutions. In situations where seeking an optimal solution is not feasible, approximation methods enable an agent to find satisfactory solutions that can be found with a reasonable amount of computation. Approximation methods can be categorized into two classes. The first class finds solutions of qualities within a predefined acceptance bound [22]. However, finding approximate solutions with a predefined quality for some difficult problems, such as evaluation of Bayesian belief networks and graph coloring, still remains NP-hard [7, 22]. In addition, algorithms in this class usually cannot generate a useful result before the end of execution. The second class of approximation methods consists of ***anytime algorithms*** [9], which first finds a solution quickly, and then successively improves the quality of the best solution at hand as long as more computation is available. Therefore, these methods do not have to set their goals in advance. The challenge for anytime algorithms is how to find good solutions as soon as possible. Many existing incremental refinement and iterative improvement methods have been used as anytime algorithms. One important anytime algorithm is local search [23]. Starting with a low-quality solution, local search repeatedly improves the current solution with local perturbations until it reaches a local minimum. It then repeats this procedure with a new starting solution if more computation is available. Another anytime algorithm is truncated depth-first branch-and-bound (DFBnB) [15, 27]. Truncated DFBnB executes DFBnB until it exhausts all available computation. The best solution found so far is then an approximate solution.

In this paper, we present two general and efficient state-space reduction methods for flexible and approximate computation. These two methods can be used to help allocate the amount of computation that is required in order to derive a decision of desired quality. Specifically, a state-space reduction is a process of reducing a state space that is difficult to search into a less complex state space that is easier to explore. In other words, a state-space reduction leads search effort to the area of problem space that is promising to provide the best approximate solution with the available computation resources. After the reduction, the optimal goal in the reduced state space, which is relatively easy to find, is found and used as an approximate solution to the original problem. By successively searching more and more complex state spaces, better solutions can be incrementally found.

The first reduction method, named as *quantitative state-space reduction*, is motivated to reduce computational complexity caused by the lack of sufficient or precise information about the problem to be solved. It treats inaccurate information as if it were more accurate in order to reduce complexity. The second reduction method, named as *structural state-space reduction*, is motivated to reduce computational complexity resulted from complex state-space structures. It abandons or postpones exploring nonpromising search avenues. Both quantitative and structural reduction methods can be applied repeatedly to incrementally provide better solutions with additional computation.

The idea of quantitative reduction was originally developed in [25, 31] for finding approximate solutions to combinatorial optimization problems. The basic idea of structural reduction can be traced back to beam search [4], an old heuristic technique for reducing search complexity. This heuristic technique was recently turned into a complete, anytime search algorithm [28]. In this paper, we re-examine these ideas and techniques for flexible and approximate computation.

The paper is organized as follows. In Section 2, we briefly discuss state-space search. In Section 3 and Section 4, we describe quantitative and structural state-space reduction methods, respectively, along with experimental results. Finally, we conclude and discuss future work in Section 5.

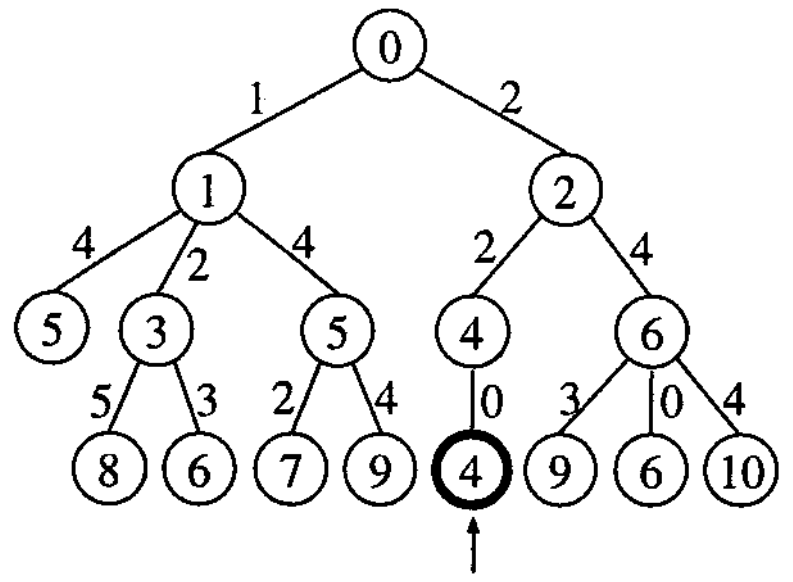

Figure 1: An incremental random tree.

## 2 State-Space Search

### 2.1 State space and search algorithms

Problem solving can be considered as a search in a state space, in which nodes represent states and edges represent operators that map one state to another. A state-space tree is a special state space which has been used extensively. In a state-space tree, a leaf node is a goal state or a state that cannot lead to a solution, A node's cost is the estimate of the actual cost of solving the problem through that node. An important class of node costs is *monotonic*, in the sense that a node cost is monotonically nondecreasing with its depth in the search tree. For most real-world problems, monotonic node costs are available or can be easily derived [24].

A state-space tree can also be viewed as if it has edge costs. The cost of an edge connecting two nodes is the cost difference between the child node and the parent, or the cost of the operator mapping the parent to the child. A state space can be captured by an abstract model called *incremental random tree*, which has been used to analyze many state-space search algorithms [17, 19, 29, 30]. This model is illustrated by Figure 1 and is defined as follows.

**Definition 2.1** *An* incremental random tree $T(b, d)$ *is a tree with depth d, variable branching factors with mean b, and non-negative variable edge costs. A node cost is the sum of the edge costs along the path from the root to that node. An optimal goal is a node of minimum cost at depth d.*

Best-first search (BFS) and depth-first branch-and-bound (DFBnB) [24], which are special cases of branch-and-bound (BnB), can be used to find an optimal goal. BFS maintains a partially expanded state space, and at each cycle expands a minimum-cost node among all those generated but not yet expanded, until an optimal goal is chosen for expansion. BFS is optimal among all algorithms that are guaranteed to find an optimal goal node using the same cost function, up to tie breaking [11]. Therefore, the complexity of BFS is the complexity of the problem, in terms of the number of nodes generated. However, BFS usually requires memory exponential in search depth, making it impractical for large problems. DFBnB uses an upper bound $u$ on the cost of an optimal goal. Starting at the root node, it chooses a most recently generated node, and then either expands this node if its cost is less than $u$, or prunes the node if its cost is greater than or equal to $u$. Whenever a new leaf node is found whose cost is less than $u$, $u$ is revised to its cost. Since DFBnB only needs memory linear in the search depth, it is preferable for large problems in practice [30].

### 2.2 Why state-space search is difficult

Two sources make a state space difficult to search. The first is the lack of sufficient or precise information about the problem to be solved. It is known that a very limited search is required if an accurate heuristic evaluation function is given. However, inaccurate heuristic functions generally prevents a problem from being solvable in polynomial time in terms of the problem size, even in an average case [24, 30].

The second source that leads to a difficult state space is the structure of a state space itself. For example, a state-space graph is more complex than a state-space tree. A concrete example is that a constraint network can be solved optimally in linear time if the network is tree structured [10], while finding solutions of a constraint network is NP-complete in general [18].

## 3 Quantitative State-Space Reduction

Quantitative state-space reduction was motivated to deal with computational complexity introduced by the lack of sufficient or precise information about the problem to be solved. The idea is to treat inaccurate heuristic information as if it were accurate, so as to speed up search process with a penalty on solution quality.

### 3.1 Phase transition of state-space search

The idea to treat an inaccurate heuristic function as an accurate one stems in a phase transition of state-space search. Consider the computational behavior of BFS and DFBnB on an incremental random tree. Let $p_0$ be the probability that an edge has a cost of zero, and $b$ be the mean branching factor. Then $bp_0$ is the expected number of children of a node whose costs are the same as that of their parent, which are referred to as *same-cost children* of the node.

**Theorem 3.1** [17, 19, 30] *On an incremental random tree $T(b,d)$, as $d \to \infty$, (1) when $bp_0 < 1$, both BFS and DFBnB generate $\theta(\gamma^d)$ nodes on average for some constant $\gamma$, $1 < \gamma \leq b$, (2) when $bp_0 = 1$, BFS generates $\theta(d^2)$ nodes and DFBnB generates $O(d^3)$ nodes on average, and (3) when $bp_0 > 1$, BFS generates $\theta(d)$ nodes and DFBnB generates $O(d^2)$ nodes on average.*

Following the optimality of BFS [11], Theorem 3.1 means that the expected complexity of finding an optimal goal experiences an abrupt transition, from exponential to polynomial in the search depth. This phenomenon is similar to a *phase transition*, which is a dramatic change to a problem property as a control parameter changes across a critical point. A simple example of a phase transition is that water changes from a solid phase to a liquid phase when the temperature rises from below the freezing point to above that point. This complexity transition is studied in great detail in [30] and is illustrated by Figure 2.

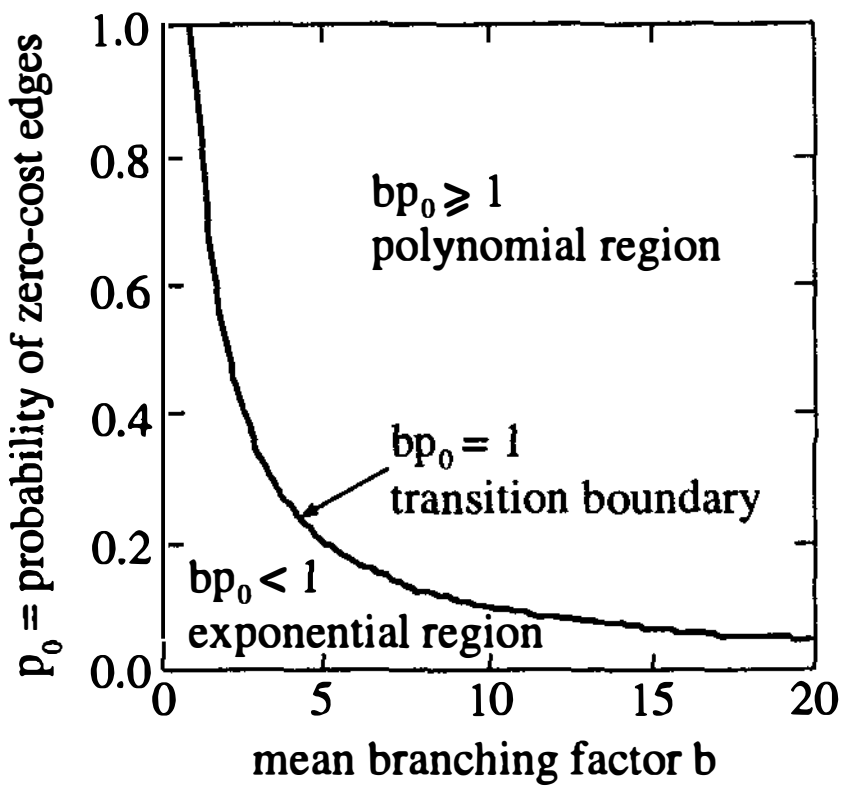

Figure 2: Complexity transition of state-space search.

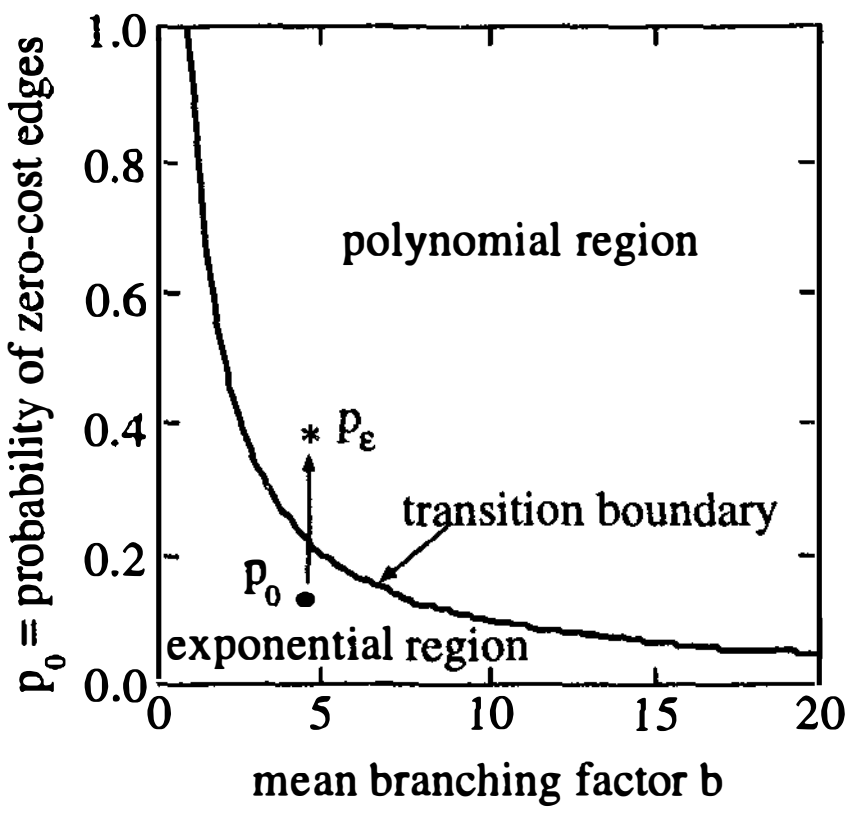

Figure 3: Reducing a difficult problem to an easy one.

### 3.2 Quantitative state-space reduction

Quantitative reduction reduces a difficult state space in the exponential region of Figure 2 into an easy one in the polynomial region. To this end, we artificially increase the expected number of same-cost children $bp_0$ of an incremental tree $T(b,d)$. This can be done by raising $p_0$, as shown in Figure 3, by artificially setting some non-zero edge costs to zero. By doing this, we also change resulting node costs accordingly.

By setting a nonzero edge cost to zero, we actually treat an inaccurate heuristic evaluation function as if it were relatively more "accurate". The accuracy of a heuristic evaluation can be measured by the cost of the optimal goal node of a state space. The smaller the cost of an optimal goal node, the more accurate the evaluation function [30]. The cost of an optimal goal node in the reduced state space is no larger than the cost of an optimal goal node in the original state space, because quantitative reduction reduces some edge costs to zero. Thus, the heuristic evaluation of the reduced state space is relatively more "accurate" than that of the original state space. In other words, this reduction gives rise to a new heuristic evaluation function with a quantitatively higher quality, thus the name quantitative reduction.

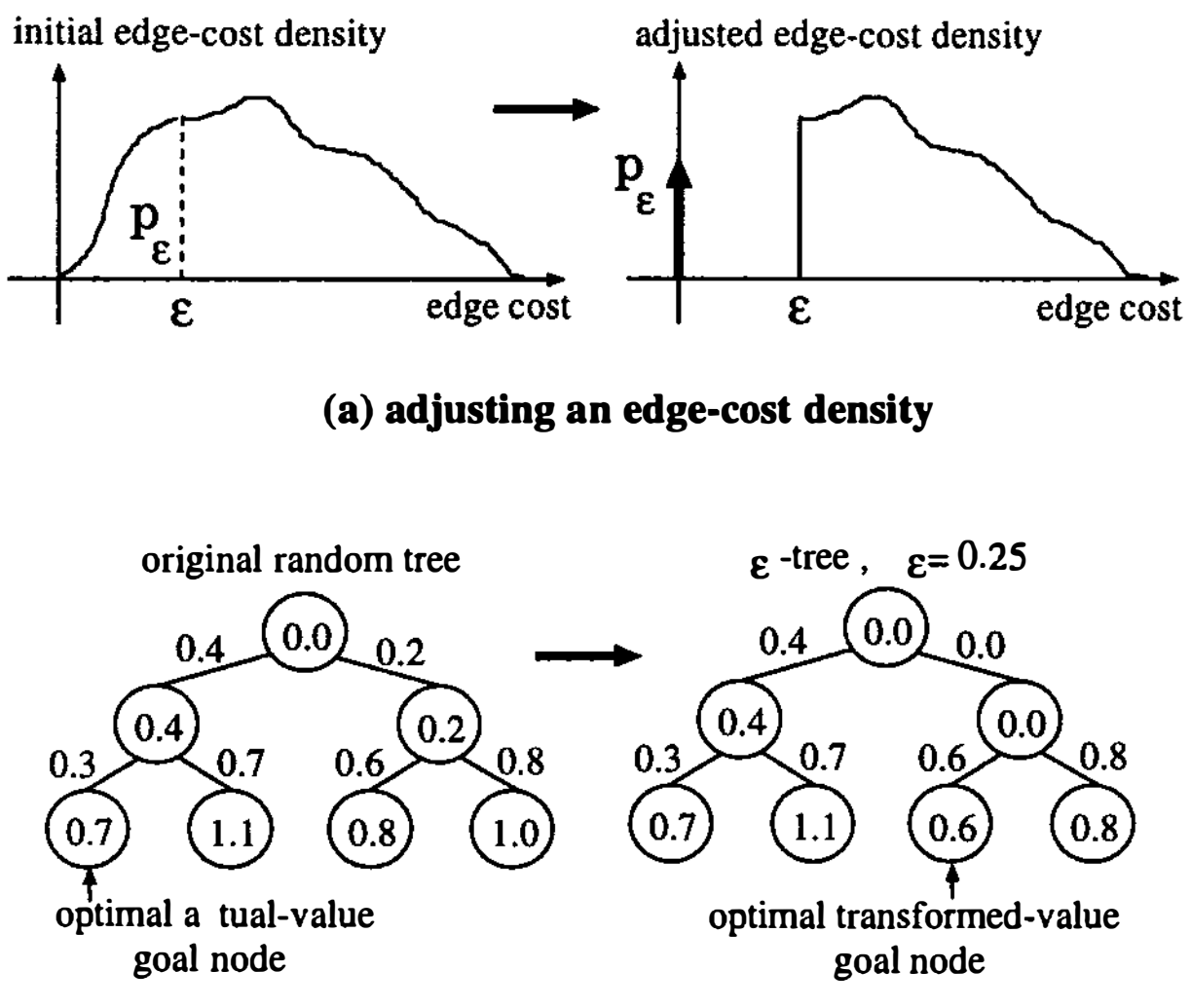

Figure 4: An example of quantitative reduction.

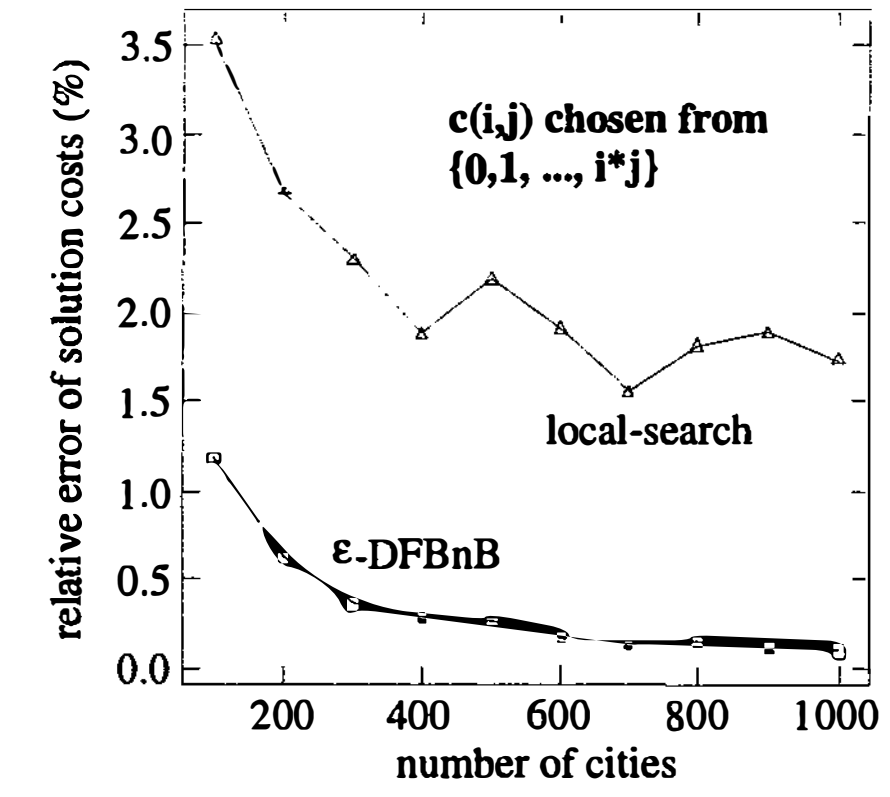

Figure 5: $\varepsilon^*$-DFBnB versus local search on the ATSP.

However, setting a nonzero edge cost to zero causes the problem of loosing heuristic information embedded in heuristic evaluation function. In order to minimize the amount of information lost and to improve the expected solution quality, we only set to zero those edge costs that are below a particular value $\varepsilon$. Given an incremental tree $T(b,d)$ and a value of $\varepsilon$, we call the tree generated by quantitative reduction an $\varepsilon$-tree $T_\varepsilon(b,d)$. An $\varepsilon$-tree is still an incremental tree, but with an adjusted edge-cost distribution, i.e., an increased probability of zero-cost edges. Figure 4(a) illustrates an edge-cost density function and its adjusted density function for a given $\varepsilon$. Figure 4(b) shows a tree $T(b = 2, d = 2)$ and its corresponding reduced tree $T_\varepsilon(b = 2, d = 2)$ with $\varepsilon = 0.25$. Moreover, the special $\varepsilon$ that reduces a state space to one on the transition boundary (see Figure 3) is called $\varepsilon^*$.

After the reduction, best-first search (BFS) or depth-first branch-and-bound (DFBnB) can be used to find an optimal goal node of the $\varepsilon$-tree, and return the actual value of this goal node in the original state space. For notational simplicity, we refer to BnB (BFS or DFBnB) using quantitative reduction as $\varepsilon$-BnB ($\varepsilon$-BFS or $\varepsilon$-DFBnB). The expected performance of $\varepsilon^*$-BnB is summarized in the following theorem.

**Theorem 3.2** [25, 31] *On an incremental tree $T(b,d)$ with $bp_0 < 1$, as $d \to \infty$, $\varepsilon^*$-BnB runs in an expected time that is at most cubic in d, and finds a goal whose expected relative solution cost error is almost surely a constant less than or equal to $(\lambda/\alpha - 1)$, where $\alpha$ is a constant, and $\lambda = E[\text{edge cost } x \mid x \le \varepsilon^*]$.*

A useful and important feature of quantitative reduction is a tradeoff between the running time of $\varepsilon$-BnB and the solution quality. Solutions with higher costs (lower quality) can be produced with less computation by using a larger value of $\varepsilon$, on average. Similarly, solutions with lower costs (higher quality) can be generated with greater computation by using a smaller value of $\varepsilon$, on average.

### 3.3 Performance of quantitative reduction

We have applied quantitative reduction to several combinatorial optimization problems, including the Traveling Salesman Problem (TSP) and constraint satisfaction problems [23]. We now report our results on the asymmetric TSP (ATSP). Given $n$ cities, $\{1, 2, 3, \cdots, n\}$, and a matrix $(c_{i,j})$ of intercity costs that defines a between each pair of cities, the TSP is to find a minimum-cost tour that visits each city exactly once and returns to the starting city. When $(c_{i,j})$ is asymmetric, i.e., the cost from city $i$ to city $j$ is not necessarily equal to that from $j$ to $i$, the problem is referred to as the ATSP.

The best cost function to the ATSP is the assignment problem [23], which is a relaxation of the ATSP since the assignments do not need to form a single tour, but instead can form a collection of disjoint subtours. If the solution to the assignment problem happens to be a single complete tour, it is also the solution to the ATSP. When the solution to the assignment problem is not a complete tour, it is decomposed, generating subproblems. In our implementation of DFBnB, we adopted Carpaneto and Toth's [6] method to generate subproblems.

To set the value of $\varepsilon^*$, we need information of node branching factors and edge costs in state space. We use an online sampling method to empirically calculate the node branching factors and distribution of the edge costs. Consider DFBnB, the algorithm we used in our implementation. DFBnB can take all nodes generated in the process of reaching the first leaf node as samples, and use them to estimate the branching factor and edge-cost distribution. These estimates are then used to calculate a value for $\varepsilon^*$. As the search proceeds, the estimates of the branching factor and edge-cost distribution can be refined and used to update the value of $\varepsilon^*$.

We compared $\varepsilon^*$-DFBnB with local search for the ATSP [16]. The average running time of local search is longer than that of $\varepsilon$-DFBnB on four different problem structures we considered. Figure 5 shows the results on the ATSP with $(c_{i,j})$ uniformly selected from $\{0, 1, \cdots, i \times j\}$, which are known to be very difficult for BnB using the assignment problem evaluation function [20]. Each data point in Figure 5 is averaged over 100 problems, ranging from 100 cities to 1,000 cities. The results show that $\varepsilon$-DFBnB outperforms local search: it runs faster and finds better solutions.

### 3.4 Iterative quantitative reduction

Quantitative reduction can be applied successively. BnB can search for better solutions with a series of quantitative reductions, with the value of $\varepsilon$ reduced after each iteration. The resulting algorithm is called iterative $\varepsilon$-BnB. Iterative $\varepsilon$-BnB can detect its termination conditions, can continuously improve the quality of the current best solution with more computation, and can ultimately find the optimal solution.

Iterative quantitative reduction can be used to construct an anytime algorithm. Combined with BFS, it turns BFS into an anytime algorithm. BFS itself does not provide a solution before its termination. By using quantitative reduction, BFS in one iteration searches a small portion of the original state space and is able to find a suboptimal goal node quickly. BFS in subsequent iterations finds better goal nodes by exploring larger portions of the state space.

Iterative quantitative reduction can also improve the anytime performance of DFBnB, which is an anytime algorithm by nature. To see this, we examined the performance profile of iterative quantitative reduction. A performance profile is usually defined by the quality of the solution found and the penalty caused by a response delay. To simplify our discussion and due to the fact that the penalty of a response delay is usually application dependent, we only consider solution quality in performance profile in the rest of this paper. The solution quality is measured by the error of a solution cost relative to the optimal cost. Denote $prof(A,t)$ as the performance profile of algorithm $A$. We define $prof(A,t)$ as

$$prof(A,t) = 1 - error(A,t), \qquad (1)$$

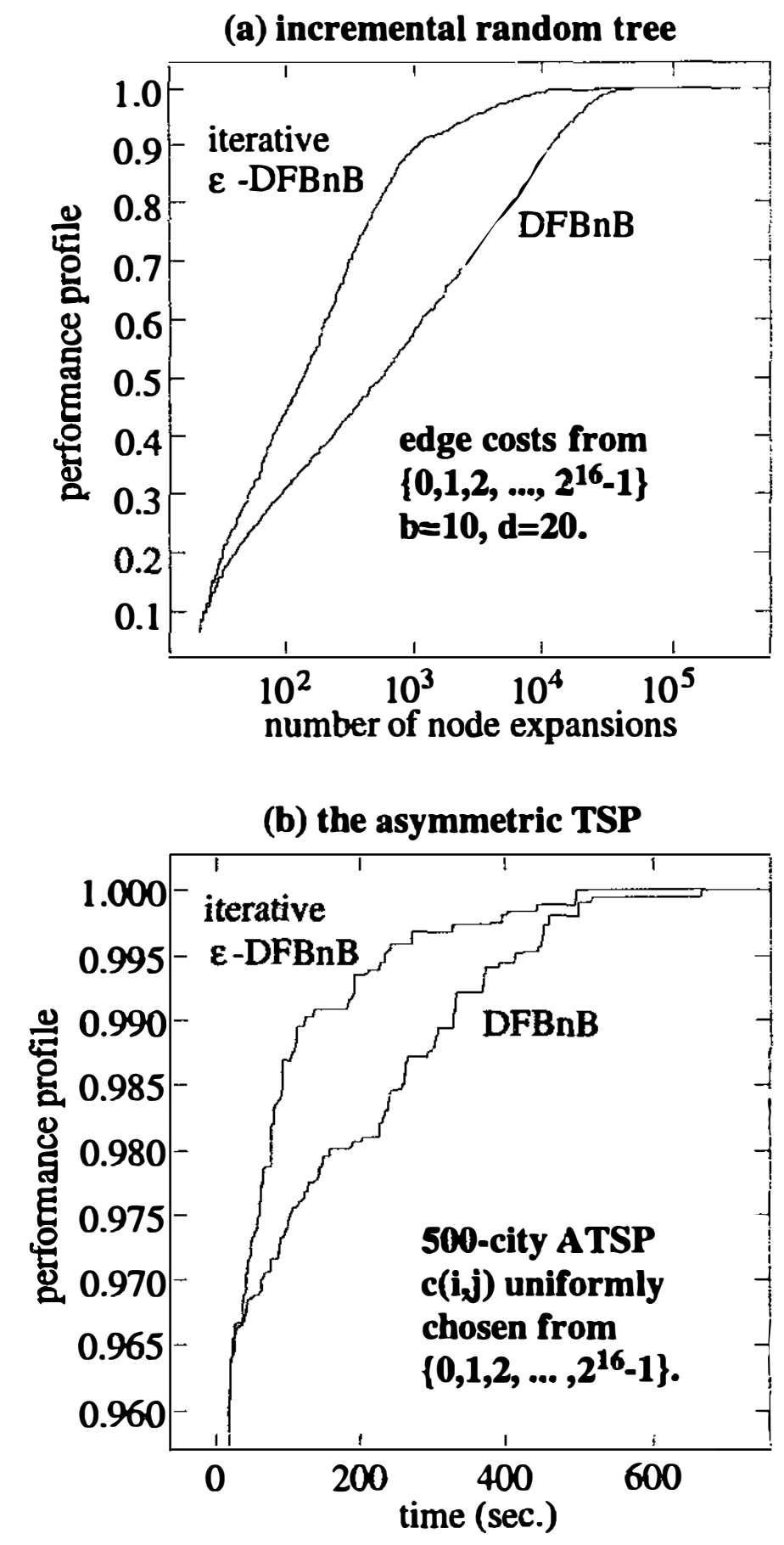

Figure 6: Iterative $\varepsilon$-DFBnB versus DFBnB.

where $error(A,t)$ is the error of solution cost of $A$ at time $t$ relative to the optimal solution cost. During the execution of $A$, $prof(A,t) \leq 1$; and at the end of its execution, $prof(A,t) = 1$.

To compare anytime performance of iterative $\varepsilon$-DFBnB with that of DFBnB, we experimented on incremental random trees and the ATSP. The first iteration of $\varepsilon$-DFBnB uses $\varepsilon^*$, whose value was learned by the online sampling method. A subsequent iteration uses $\varepsilon$ whose value is half of that used in the previous iteration. Figure 6(a) is the result on incremental random trees $T(b = 10, d = 20)$ with edge costs uniformly chosen from $\{0, 1, 2, \cdots, 2^{16} - 1\}$. The result is averaged over 1000 instances. Figure 6(b) shows the result on 500-city ATSP's, averaged over 100 random problem instances. The horizontal axis is the CPU time on a Sun4 sparc460 workstation. Figure 6 shows that iterative $\varepsilon$-DFBnB is superior over DFBnB, finding better solutions sooner on average.

## 4 Structural State-Space Reduction

As mentioned in Section 2, the second source that makes a state space difficult to search is the state-space structure itself. Structural state-space reduction was motivated to simplify a complex state-space structure so that high quality solutions can be found quickly.

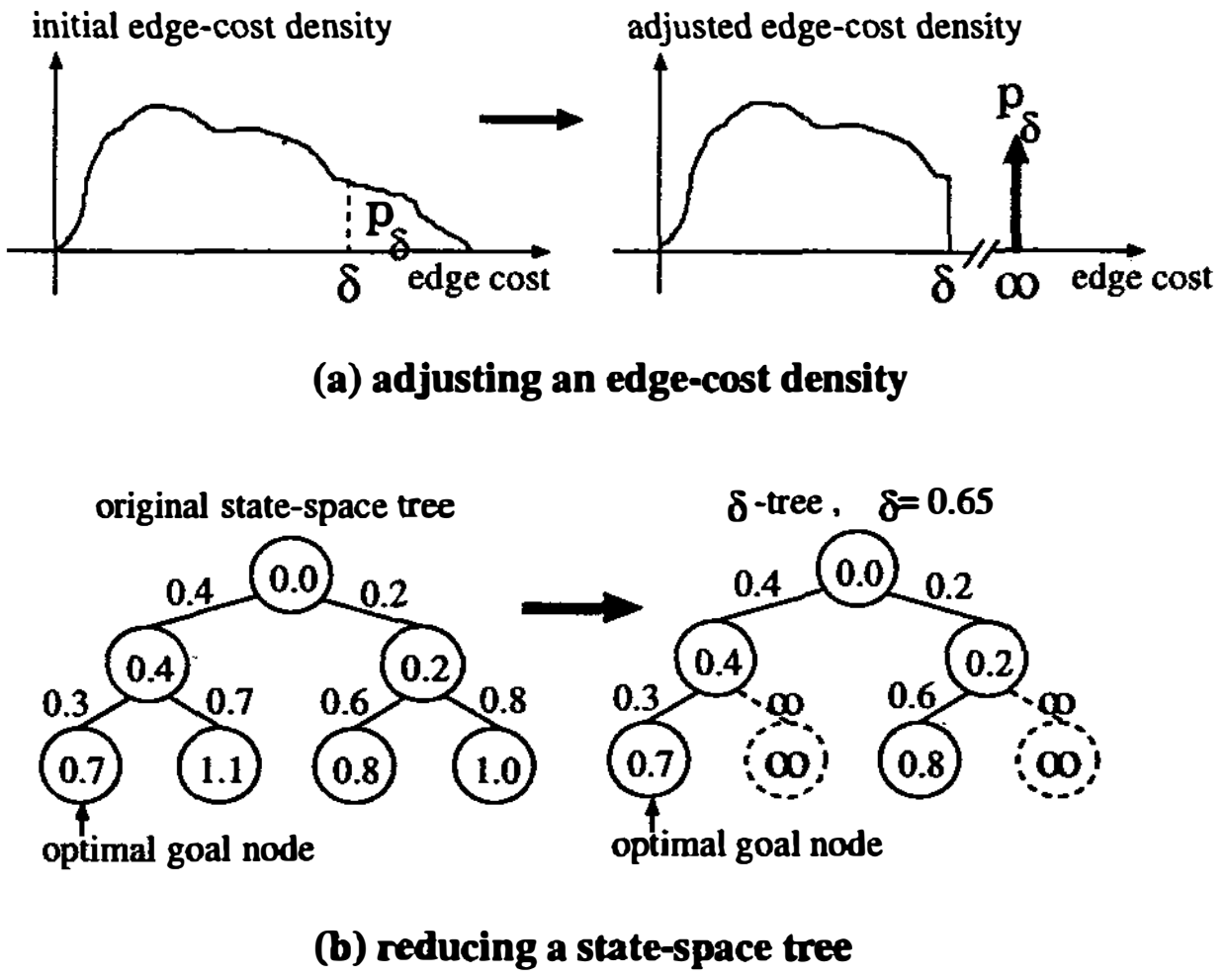

Figure 7: An example of structural reduction.

### 4.1 Structural state-space reduction

To simplify our discussion, we only consider state-space tree, since a state-space graph can be represented by state-space tree. Consider again an incremental search tree. Intuitively, a large edge cost is more likely to lead to large costs of nodes generated below this edge, since a node cost is the sum of the edge costs leading to it. Symmetric to quantitative reduction, we can artificially set edge costs that are greater than a parameter $\delta$ to infinity. By doing this, we actually prune some search avenues, so as to generate a simplified state space with fewer nodes. This reduction may change the structure of a state space being explored, thus the name ***structural state-space reduction***.

After a structural reduction, BFS or DFBnB can be used to find an optimal goal node of the reduced state space and return this goal node and its cost as the result. BFS or DFBnB using structural reduction is referred to as $\delta$-BFS or $\delta$-DFBnB, respectively.

Compared to the original state space, the reduced one has an adjusted edge-cost distribution. Figure 7(a) shows an edge-cost density function and its adjusted density function for a given $\delta$. we call a tree reduced from an incremental tree a $\delta$-tree $T_\delta$. Figure 7(b) shows an incremental tree $T(b = 2, d = 2)$ and its corresponding $\delta$-tree $T_\delta(b = 2, d = 2)$ with $\delta = 0.65$.

However, structural reduction is a two-edged sword. On one side, it reduces the amount of search by reducing node branching factors. On the other side, it runs the risk of missing a goal. The reason that structural reduction may not find a goal node at all is that it abandons too many search alternatives and create deadend nodes, the ones that do not have a child node. It turns out that the deadend nodes in state space have a great impact on the possibility that structural reduction can find a goal. A direct, but partial remedy to the failure of reaching a goal node is to keep one child node when all children of a node were to be pruned. This can greatly increase the possibility that structural reduction find a solution. To find a high quality goal node, the child node with the minimum cost among all the children of a node should be kept.

### 4.2 Iterative structural reduction

The biggest drawback of structural reduction is that it may fail to find a solution even if one exists. There are also situations where the solution found by structural reduction has a low quality. To overcome these difficulties, we can apply structural reduction in iterations, using a larger value of $\delta$ after each iteration. This allows us to explore increasingly larger and more complex state spaces in order to find better solutions. Iterative structural reduction terminates under one of the following two situations. First, when a satisfied goal is found, it can quit. Second, if no pruning from structural reduction has been applied in an iteration, the algorithm can terminate with an optimal goal. This is because the last iteration runs the underlying search method with no extra pruning from structural reduction, and thus finds an optimal goal node.

Similar to iterative quantitative reduction, iterative structural reduction can turn BFS into an anytime algorithm. In the rest of this section, we compare the anytime performance of iterative $\delta$-DFBnB to that of DFBnB, using performance profile of (1).

We used online sampling method to learn the edge cost distribution and to compute the value of $\delta$. With this sampling method, the first iteration of structural reduction does not abandon a node until a certain number of nodes have been generated. In our implementation, structural reduction does not prune a node until it has reached the first leaf node. All the nodes generated in the process of reaching the first leaf node are used as initial samples for computing an empirical distribution. Furthermore, using the empirical distribution of node-cost differences, we can set parameter $\delta$. In our experiments, the initial $\delta$ is set to a value $\delta_I$ such that a node-cost difference is less than $\delta_I$ with probability $p$ equal to 0.1. The next iteration increases probability $p$ by 0.1, and so forth, until no reduction has been applied in the latest iteration or probability $p$ is greater than 1.

We compared $\delta$-DFBnB against DFBnB on maximum 3-satisfiability (3-Sat) [12] using a variation of Davis-Putnam method [8]. We generated maximum 3-Sat problem instances by randomly selecting three variables and negating them with probability 0.5 for each clause. Duplicate clauses were removed. The problem instances we used have a large ratio of the number of clauses to the number of variables (clause to variable ratio), since random 3-Sat problems with a small clause-to-variable ratio are generally satisfiable [21]. Figure 8(a) shows the experimental result of on 3-Sat with 30 variables and 450 clauses, averaged over 100 random problem instances. Figure 8(a) shows that iterative $\delta$-DFBnB significantly improves the anytime performance of DFBnB, finding better solutions sooner. For instance, with total of 20,000 node generations, the average error of solution found relative to the optimal is 4.1% (profile=0.959) from iterative $\delta$-DFBnB, while the average error is 15.9% (profile=0.841) from DFBnB.

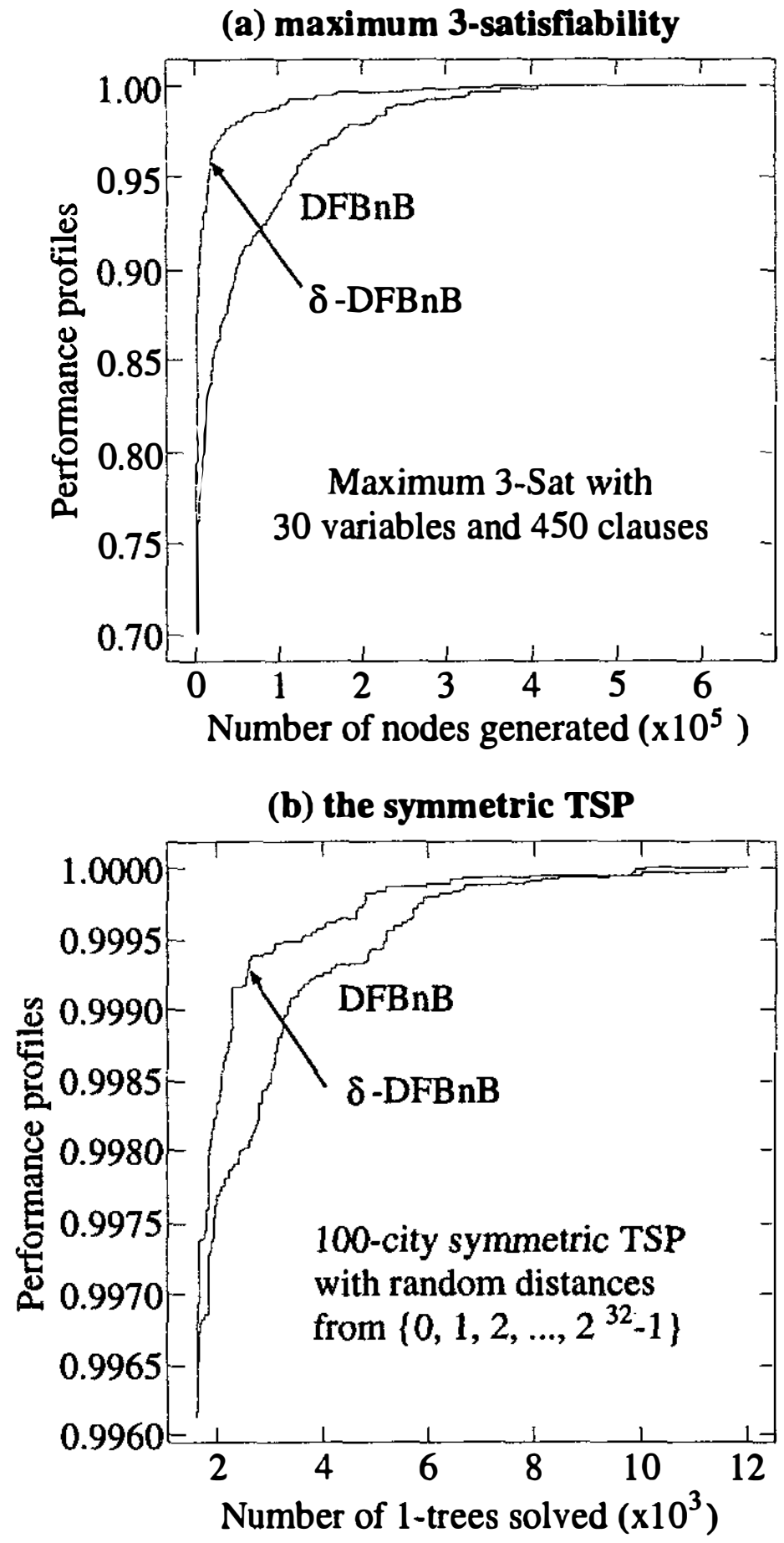

Figure 8: Iterative $\delta$-DFBnB versus DFBnB.

We also compared iterative $\delta$-DFBnB against DFBnB on the symmetric TSP (STSP), in which the cost from city $i$ to city $j$ is the same as that of from $j$ to $i$. In our implementation, we use Held-Karp lower bound function [13] to compute node costs. This cost function iteratively computes a Lagrangian relaxation on the STSP, with each step constructing a 1-tree. A 1-tree is a minimum spanning tree (MST) [23] on $n-1$ cities plus the two shortest edges from the city not in the MST to two cities in the MST. Note that a complete TSP tour is a 1-tree. If no complete TSP tour has been found after a predefined number of steps of Lagrangian relaxation, which is $n/2$ in our experiment, the problem is decomposed into at most three subproblems using the Volgenant and Jonker's branching rule [26]. We generated STSP problem instances by uniformly choosing a cost between two cities from $\{0, 1, 2, \cdots, 2^{32}-1\}$. Figure 8(b) shows the experimental result on 100-city random STSPs, averaged over 100 instances. It shows that the anytime performance of iterative $\delta$-DFBnB is superior to that of DFBnB.

## 5 Conclusions and Future Work

We have presented two state-space reduction methods for flexible and approximate computation. The main idea is to reduce a state space that is difficult to search into a state space that is easy to explore. After a reduction, the optimal goal in the reduced state space is found and used as an approximate solution to the original problem. Specifically, we have developed quantitative state-space reduction to treat inaccurate information as if it were more accurate to reduce computational complexity, and structural state-space reduction to deal with complex state space structures by abandoning or postponing exploration of some non-promising search avenues. We have also described how these two methods can be applied repeatedly to incrementally find better solutions with additional computation. Our experimental results show that the state-space reduction techniques are (a) general, which can be applied to different problem domains due to the generality of state-space representation; (b) effective for approximation, finding better solution with less amount of computation than some existing approximation methods on combinatorial optimization problems such as maximum 3-Satisfiability and the Traveling Salesman Problem; (c) and efficient for flexible computation, which provides a means to trade computation time for solution quality and to continuously improve solution quality with additional computation.

The current version of structural reduction is relatively rigid, in the sense that it will prune a branch if its cost is greater than a predefined value $\delta$. Alternatively, we may set edge costs that are larger than $\delta$ to a large, non-infinity value to postpone the exploration of the subtrees underneath the edges.

Quantitative reduction and structural reduction are orthogonal and complementary. We are currently combining these two reduction methods to deal with high computational complexity caused by both inaccurate heuristic information and complex problem structures.

## Acknowledgments

This work was partially funded by NSF Grant IRI-9619554. Thanks to Joe Pemberton and Rich Korf for collaboration and discussions related to heuristic search and phase transitions.